\documentclass[twoside,leqno,twocolumn]{article}
\usepackage{ltexpprt}

\usepackage{amssymb}
\usepackage{amsmath}

\usepackage{graphicx}

\usepackage{times}

\usepackage{graphicx}

\usepackage[caption=false]{subfig}

\usepackage{float}
\usepackage{url}
\usepackage{enumitem}
\usepackage{pgfplots}

\usepackage{tikz}
\usetikzlibrary{positioning}

\usepackage{hhline}
\usepackage{multirow}
\usepackage{booktabs}
\usepackage{wrapfig}
\usepackage{algorithm}
\usepackage{algpseudocode}

\newcommand{\eat}[1]{}

\usepackage{pifont}

\usepackage{tikz}
\usetikzlibrary{arrows}
\usetikzlibrary{fit}
\usetikzlibrary{positioning,shapes}
\usetikzlibrary{shapes.misc,matrix,fit,calc,patterns}

\usepackage{url}

\begin{document}
\title{DiSMEC - Distributed Sparse Machines for Extreme Multi-label Classification}
\author{Rohit Babbar and Bernhard Sch\"{o}lkopf\\
 MPI for Intelligent Systems}
\maketitle
\begin{abstract}
Extreme multi-label classification refers to supervised multi-label learning involving hundreds of thousands or even millions of labels. Datasets in extreme classification exhibit fit to power-law distribution, i.e. a large fraction of labels have very few positive instances in the data distribution. Most state-of-the-art approaches for extreme multi-label classification attempt to capture correlation among labels by embedding the label matrix to a low-dimensional linear sub-space. However, in the presence of power-law distributed extremely large and diverse label spaces, structural assumptions such as low rank can be easily violated.

In this work, we present DiSMEC, which is a large-scale distributed framework for learning one-versus-rest linear classifiers coupled with explicit capacity control to control model size. Unlike most state-of-the-art methods, DiSMEC does not make any low rank assumptions on the label matrix. Using double layer of parallelization, DiSMEC can learn classifiers for datasets consisting hundreds of thousands labels within few hours. The explicit capacity control mechanism filters out spurious parameters which keep the model compact in size, without losing prediction accuracy. We conduct extensive empirical evaluation on publicly available real-world datasets consisting upto 670,000 labels. We compare DiSMEC with recent state-of-the-art approaches, including - SLEEC which is a leading approach for learning sparse local embeddings, and FastXML which is a tree-based approach optimizing ranking based loss function. On some of the datasets, DiSMEC can significantly boost prediction accuracies - 10\% better compared to SLECC and 15\% better compared to FastXML, in absolute terms. 
\end{abstract}

\section{Introduction}
With the emergence of big data, large-scale classification problems have gained considerable momentum in recent years.
In this respect, supervised learning with a large target label set has attracted attention of machine learning researchers and practitioners.
Datasets consisting of hundreds of thousand possible labels are common in various domains such as product categorization for e-commerce ~\cite{mcauley2013hidden,shen2011item, Bengio10, Agrawal13}, large-scale classification of images ~\cite{deng2010does, krizhevsky2012imagenet} and text ~\cite{gopal2013recursive,  Prabhu14}.

In this work, we focus on Extreme multi-label classification (\textbf{XMC}), which refers to multi-label classification where the label set has dimensionality of the order of hundreds of thousands or even millions.
The goal in XMC is to learn a classifier which can annotate an unseen instance with a relevant subset of labels from the extremely large set of all possible labels.
For instance, all Wikipedia pages are tagged with a small set of relevant labels which are chosen from a large set consisting of more than a million possible tags in the collection.
XMC framework can be employed to build a classifier which can automatically label hundreds of new pages generated in Wikipedia everyday.
Similarly, in image annotation tasks, one might want to build a classifier which can automatically tag individuals appearing in a photo from the set of all possible people tags in the repository.
Furthermore, the framework of XMC can leveraged to effectively address machine learning problems arising in web-scale data mining, such as recommendation systems, ranking and bid phrase suggestion for Ad-landing pages \cite{Agrawal13, prabhu2014fastxml}.
Learning from a user's buying patterns or browsing patterns in the past, this framework can be used to recommend a small subset of relevant items or personalized search results from an extremely large set of all possible items or search results.
The growing significance of XMC to tackle large-scale machine learning problems in web-scale data mining is further highlighted by dedicated workshops in premier machine learning and data mining conferences (cf. workshops on extreme classification at NIPS 2013--2015\footnote{\tiny\\
\url{http://research.microsoft.com/en-us/um/people/manik/events/xc13/}\\
\url{http://research.microsoft.com/en-us/um/people/manik/events/xc14/}\\
\url{http://research.microsoft.com/en-us/um/people/manik/events/xc15/}\\
} and WSDM 2014\footnote{\tiny\\\url{http://lshtc.iit.demokritos.gr/WSDM_WS}}).

\begin{figure*}[ht]
\centering

\subfloat[Amazon-670K dataset]{%
  \includegraphics[width=0.4\textwidth]{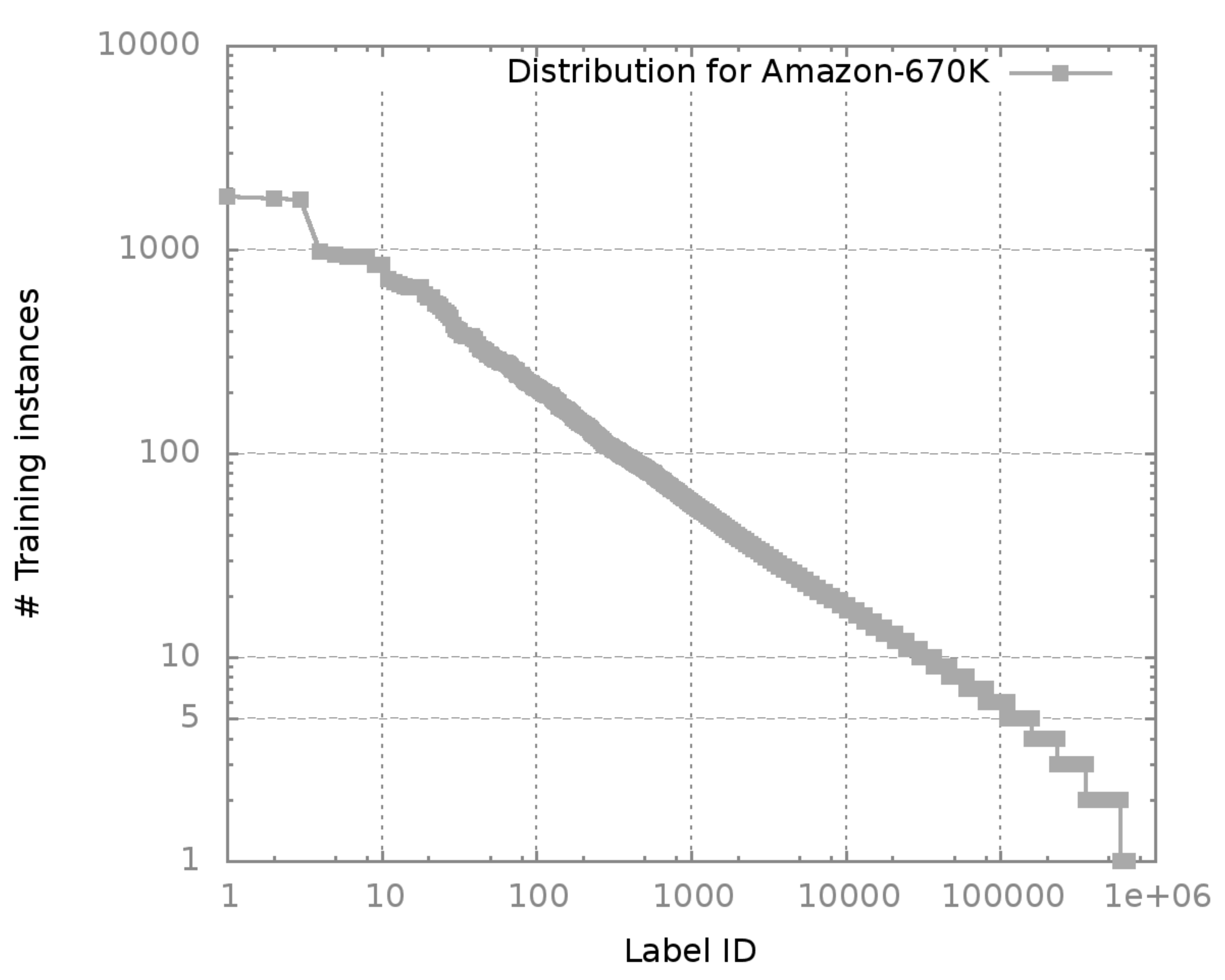}%
  \label{pl:amazon}%
}
~
\subfloat[Wikipedia-31K dataset]{%
  \includegraphics[width=0.4\textwidth]{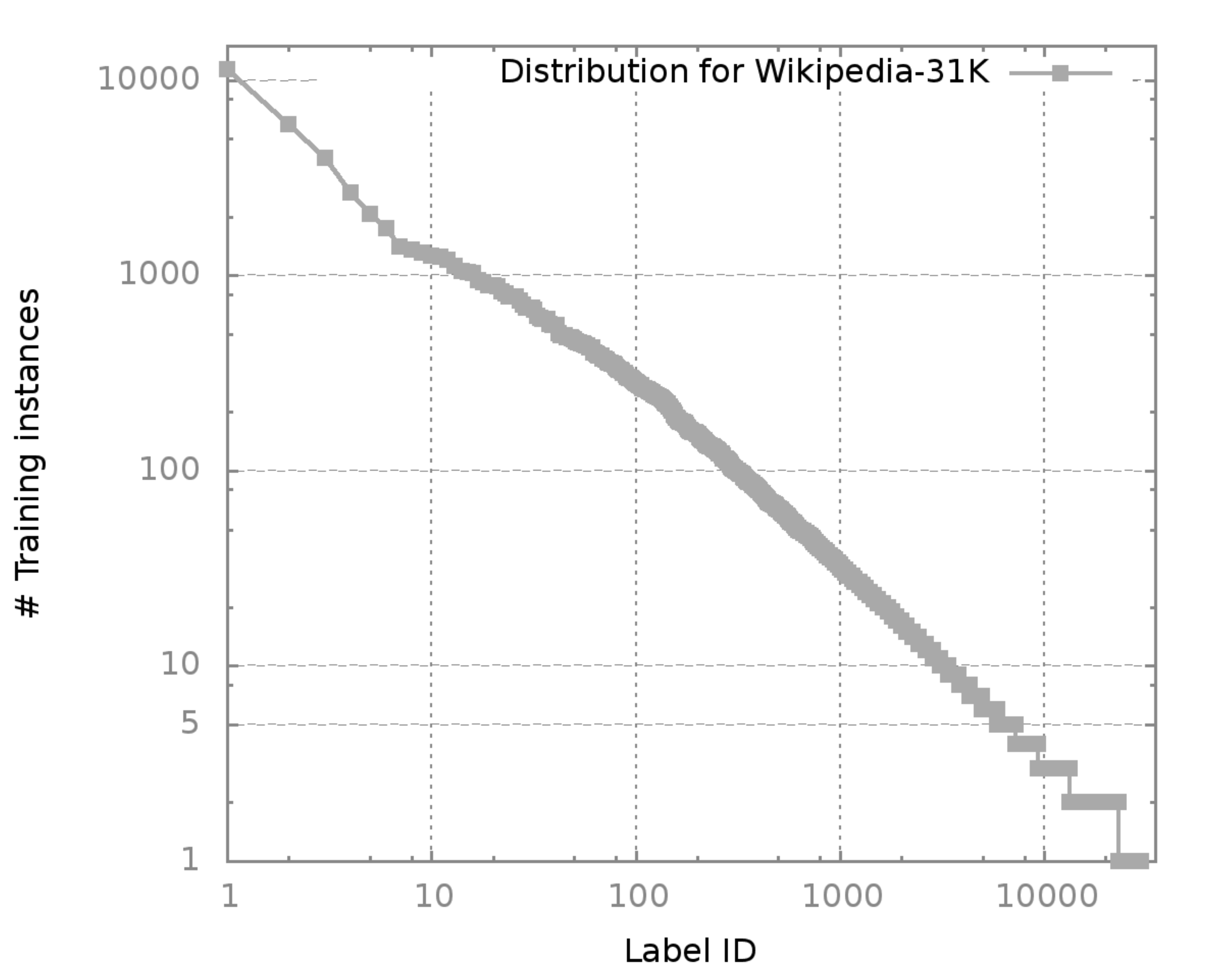}%
  \label{pl:wiki}%
}
\caption{Power-law distribution for two datasets (a) Amazon-670K, (b)Wikipedia-31K from the extreme classification repository. Out of 670,000 labels in the Amazon-670K, only 100,000 have more than 5 training instances. Similar phenomena is observed for other datasets also.}
\end{figure*}

Classical machine learning methods cannot be trivially applied to the computational and statistical challenges arising in extreme multi-label classification.
From the computational viewpoint, in addition to millions of labels, one needs to deal with millions of training instances which live in high-dimensional feature spaces.
A direct application of off-the-shelf solvers such as Liblinear requires several weeks of training and problems of storing data and model in the main memory during training. Even if one could successfully learn the classifiers, the resulting models have size of the order of hundreds of GBs or sometimes a few TBs, which becomes impossible to load to make any meaningful predictions in an efficient manner.

An important characteristic of the datasets in the XMC setting is that the distribution of training instances among labels exhibits fit to power-law distribution. 
This implies that a large fraction of labels have very few training instances assigned to them.
For Amazon-670K and Wikipedia-31K datasets, used in our experiments, which  consist of 670,000 and 31,000 labels respectively, the distribution of training instance among labels is shown in Figure 1.
Formally, let $N_r$ denote the size of the $r$-th ranked label, when ranked in decreasing order of number of training instances that belong to that label, then :
\begin{equation}
N_r = N_1r^{-\beta}
\label{eq:powerlaw1}
\end{equation} 
where $N_1$ represents the size of the 1-st ranked label and $\beta>0$ denotes the exponent of the power law distribution.
For instance, as shown in Figure \ref{pl:amazon}, only 100,000 out of 670,000 labels have more than 5 training instances that belong to them. 
From a statistical viewpoint, when a significant fraction of labels are \textit{tail labels}, it becomes hard for machine learning algorithms to learn good classifiers with very little data.
The role of fat-tailed distributions in multi-class settings and hierarchical classification has also been discussed in the recent works \cite{babbar2013flat, babbar2014re, babbar2014power}.

The XMC framework, even though quite promising in addressing practical challenges in web-scale data mining such as tagging, recommendation systems and ranking, still needs to tackle the computational, storage, and statistical challenges for its successful application.
In recent years, various approaches have been proposed to address them, which can be broadly divided into embedding-based and tree-based methods.
Both these types of approaches aim to reduce the effective space of labels in order to control the complexity of the learning problem.
\subsection{Label-embedding approaches}

Label-embedding approaches assume that the label matrix has effectively low rank and hence project it to a low-dimensional linear sub-space.
These approaches have been at the fore-front in multi-label classification for small scale problems which consist of few hundred labels \cite{bhatia2015sparse, hsu2009multi, cisse2013robust, tai2012multilabel, bi2013efficient, zhangmulti, chen2012feature, weston2011wsabie, linmulti}.
In addition to being able to capture the label correlation, these methods also exhibit strong generalization guarantees.

In XMC setting which consists of a diverse power-law distributed label space, the crucial assumption made by the embedding-based approaches of a low rank label space breaks down \cite{xurobust, bhatia2015sparse}.
Under this condition, global embedding based approaches leads to high prediction error.
Furthermore, these approaches can be slow for training and prediction in the XMC scenarios consisting of hundreds of thousand labels.

In order to overcome these limitations, recently proposed SLEEC (Sparse Local Embedding for Extreme Classification \cite{bhatia2015sparse}), first clusters the data into smaller regions. It then performs local embeddings of label vectors by preserving distances to nearest label vectors which are learnt using a k-nearest neighbor classifier. Label matrix compression is then achieved by learning non-linear embeddings instead of linear mappings as in previous approaches.
Since clustering can be potentially unstable in extremely high dimensional spaces, an ensemble of SLEEC learners is employed to achieve good prediction accuracy. Despite its limited success, SLEEC has the following shortcomings : (i) firstly, in the presence of \textit{tail labels} where the low rank assumption is violated, it is unclear to what extent the \textit{locally} low rank assumption made by SLEEC is valid, (ii) secondly, as mentioned by the authors, the clustering step for achieving scalability to large training data sizes is an approximation and potentially unstable in high dimensions, (iii) thirdly, SLEEC obtains a local-minima while solving the non-convex rank constraint embedding problem, and hence the solution can be highly sub-optimal, and (iv) finally, SLEEC has eight hyper-parameters many of which are set arbitrarily such as - number of SLEEC learners in the ensemble are set to 5,10 or 15; number of clusters set to $\lfloor(\text{\# training points}/6000)\rfloor$; number of nearest neighbors which is set to 100 for large datasets and 50 for small datasets.

\subsection{Tree-based approaches}
Tree-based approaches are aimed towards faster prediction which can be achieved by recursively dividing the space of labels or features.
Due to the cascading effect, the prediction error made at a top-level cannot be corrected at lower levels. As a result, these methods have much lower prediction accuracy. Typically, such techniques trade-off prediction accuracy for logarithmic prediction speed which might be desired in some applications.

FastXML is a state-of-the art classifier in extreme multi-label classification, which optimizes an nDCG based ranking loss function \cite{prabhu2014fastxml}. It recursively partitions the feature space instead of the label space and uses the observation that only small number of labels are active in each region of feature space.
Partitioning of the label space for achieving sub-linear ranking has been proposed in LPSR method which uses Gini index as a measure of performance \cite{weston2013label}. This is not suited for XMC scenarios where it is more important to predict a small set of relevant labels than predicting a large number of non-relevant ones.

\subsection{Naive one-vs-rest approaches}
One-vs-rest method, learns a weight vector for each label to distinguish that label from the rest.
In recent studies \cite{bhatia2015sparse, prabhu2014fastxml, yenpd}, this technique has been ignored mainly due to the following reasons:
\begin{itemize}

\item Training Complexity - Training one-vs-rest algorithms for XMC problems involving hundreds of thousand labels using off-the-shelf solvers such as Liblinear can be computationally and memory intensive.
For WikiLSHTC-325K dataset which consists of 325,000 labels, an adhoc application of Liblinear can take 96 days to train (Table 2 in \cite{yenpd}).

\item Model size and prediction speed - It has also been observed in recent studies that models learnt for XMC datasets can be extremely large. For WikiLSHTC-325K dataset, the model learnt by using $\ell_2$ regularization in a linear SVM can have a size 870 GB (Table 2 in \cite{yenpd}). 
Furthermore, large model sizes lead to latency in prediction while evaluating the dot products of weight vectors of individual labels against the test instance.
\end{itemize}

\subsection{Our Contributions}
The above mentioned concerns about the scalability of one-vs-rest method for XMC problems are primarily due to the direct usage of off-the-shelf solvers.
In this work, we attempt to dispel the above notions by re-visiting the one-vs-rest paradigm in the context of XMC. We present a distributed learning mechanism, DiSMEC - \textbf{D}istributed \textbf{S}parse \textbf{M}achines for \textbf{E}xtreme Multi-label \textbf{C}lassification. DiSMEC can easily scale upto hundreds of thousands labels, provides significant improvements over state-of-the-art prediction accuracies, learns compact models and performs real-time prediction.
Concretely, our contributions are the following:

\begin{enumerate}
\item Doubly parallel training - DiSMEC employs a double layer of parallelization, and by exploiting as many cores as available, it can gain significant training speedup over SLEEC and other state-of-the-art methods.
On WikiLSHTC-325K, our method can learn the model in 6 hours on 400 cores, unlike single layer of parallelization \cite{yenpd}.

\item Prediction performance - In addition to computational benefits, DiSMEC achieves significant improvement in prediction performance on public benchmark datasets. On three out of seven datasets, DiSMEC achieves 10\% improvement in precision and nDCG measures over SLEEC and 15\% improvement over FastXML in absolute terms. Both SLEEC and FastXML are state-of-the-art approaches which have been employed in industrial production systems for bid-phrase recommendation and web advertising.

\item Model size and prediction speed- By explicitly inducing sparsity via pruning of spurious weights, models learnt by DiSMEC can have upto three orders of magnitude smaller size.
For WikiLSHTC-325K, DiSMEC model size is 3 GB as compared to 870 GB \cite{yenpd}. Furthermore, SLEEC's model takes 10 GB for learning an ensemble of 15 learners, and still under-performs compared to DiSMEC by 10\%. Compact models learnt by DiSMEC coupled with distributed storage of learnt models further results in faster prediction through parallel evaluation of prediction step. 

\end{enumerate}

To the best of our knowledge, our work is the first attempt for scaling up the well-known one-vs-rest paradigm for extreme multi-label classification problems.

\section{Proposed Method}

\begin{figure*}[ht]
\centering

\subfloat[Distribution of weights before pruning]{%
  \includegraphics[width=0.4\textwidth]{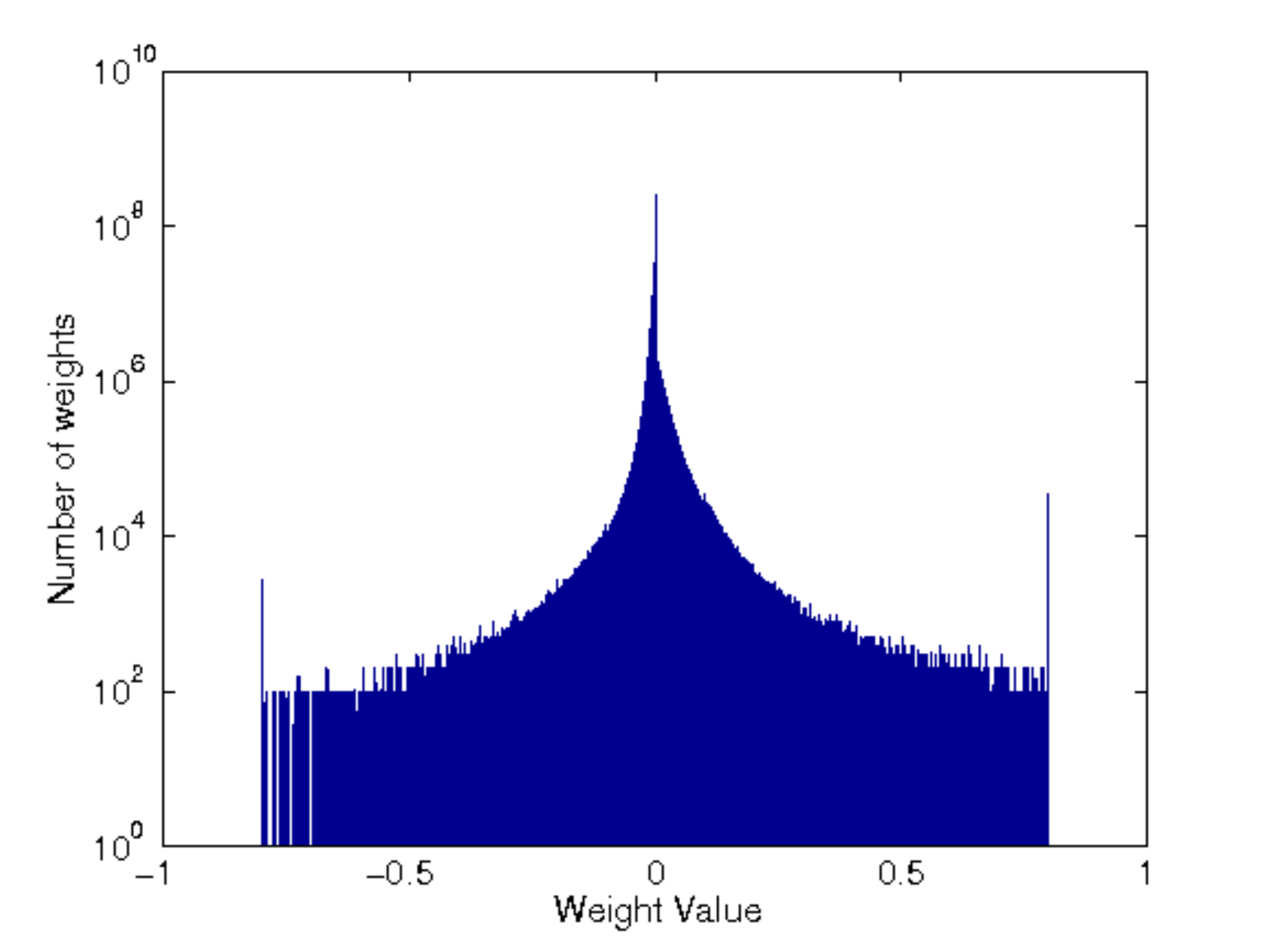}%
  \label{fig:distorg}%
}
~
\subfloat[Distribution of weights after pruning]{%
  \includegraphics[width=0.4\textwidth]{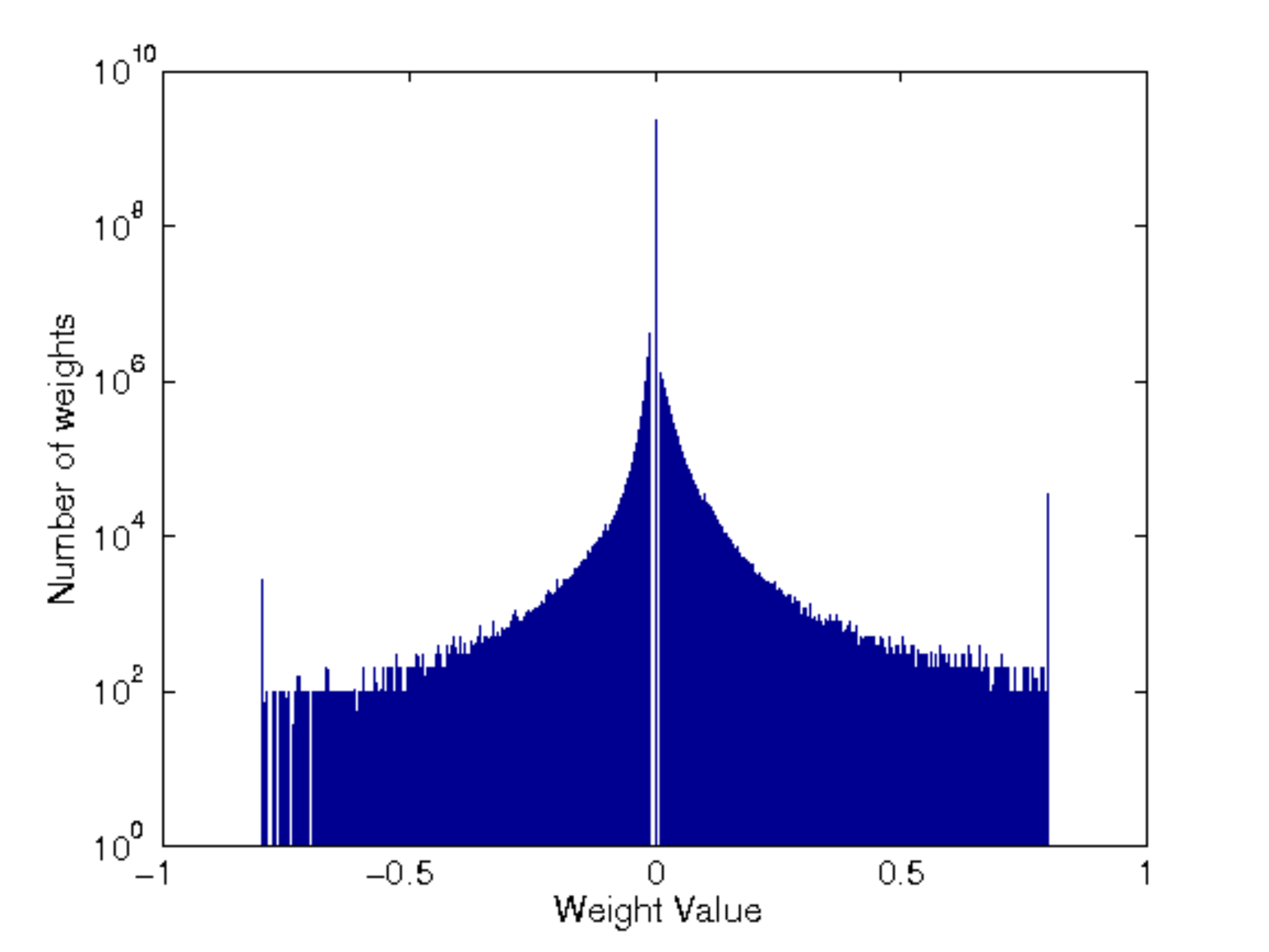}%
  \label{fig:distprune}%
}
\caption{Histogram plot depicting the distribution of learnt weights for Wikipedia-31K dataset (a) before pruning, and (b) after pruning (step 7 in Algorithm \ref{algorithm:alg1}). The shaded area in (b) consists of only 4\% of the weights as against 96\% \textit{ambiguous} weights in the region in the neighborhood of 0 which are removed after the pruning step leading to drastic reduction in model size.}
\end{figure*}

Let the training data, given by $\mathcal{T} = \{(\textbf{x}_1,\textbf{y}_1) \ldots (\textbf{x}_N,\textbf{y}_N) \}$ consist of input feature vectors $\textbf{x}_i \in \mathcal{X} \subseteq \mathbb{R}^D $ and respective output vectors $\textbf{y}_i \in \mathcal{Y} \subseteq \{0,1\}^L$ such that $y_{ij}=1$ iff the $j$-th label belongs to the training instance $\textbf{x}_i$.
The training data can be equivalently re-written in the format $\mathcal{T} = \{ \textbf{X}, \textbf{Y} \}$, where $\textbf{X} \in \mathbb{R}^{N \times D}$, such that the $i$-th row of $\textbf{X}$ consists of the $i$-th input vector, i.e., $\textbf{X}_{i,:} = \textbf{x}_i^T$. Similarly, $\textbf{Y} \in \{0,1\}^{N \times L}$, such that $i$-th row of $\textbf{Y}$ consists of the $i$-th label vector, i.e., $\textbf{Y}_{i,:} = \textbf{y}_i^T$.

In XMC setting, the training set size $N$, the feature set dimensionality $D$ and the label set size $L$ are extremely large. 
Concretely, we deal with datasets in which $N$ varies from $10^4$ to $10^6$, $D$ varies from $10^5$ to $10^6$, and $L$ varies from $10^4$ to $10^5$. 
Furthermore, both the feature vectors and the label vectors are sparse  i.e., only very few features and labels would be active for each input output pair $(\textbf{x}_i,\textbf{y}_i)$. 
The goal in XMC setting is to learn a multi-label classifier in the form of a vector-valued output function $f : \mathbb{R}^D \mapsto \{0,1\}^L$.
For high-dimensional datasets, such as those encountered in XMC settings, linear classifiers are more efficient to train and have been shown to perform competitive to non-linear methods \cite{yuan2012recent}.
The desired linear function $f$ is therefore parametrized by $L$ weight vectors $(\textbf{w}_{1}\ldots\textbf{w}_{L})$, which are learnt from the training data.
For prediction of a test instance $\textbf{x}$, the inner product is evaluated for all the weight vectors, and top ranked labels are predicted.
It has been shown that the above formal setup for XMC can be applied to address the problems such as recommending items to users based on their profile and suggestion of bid phrase to advertizers for Ad-landing pages \cite{Agrawal13}.

We propose to employ the binary one-vs-rest framework for learning the weight vector, $\textbf{w}_{\ell}$, corresponding to each label $\ell$ \cite{rifkin2004defense}. 
To learn $\textbf{w}_{\ell}$, we require input data in the format $\mathcal{T}_l = \{ \textbf{X}, \textbf{s}_l \}$, where $\textbf{s}_l \in \{+1, -1\}^N$ represents the vector of signs depending on whether $\textbf{y}_{i\ell} = 1$ or not.
Using the training data $\mathcal{T}_l$, learning the linear separator $\textbf{w}_{\ell}$ can then be achieved by solving the following optimization problem which uses squared-hinge loss and $l_2$-regularization:
\begin{equation}
\min_{\textbf{w}_{\ell}}  \left[ ||\textbf{w}_{\ell}||_2^2 + C \sum_{i=1}^{N} (\max(0,1-s_{\ell_i}\textbf{w}^T_{\ell}\textbf{x}_i))^2\right]
\end{equation}\label{objective}
where $C$ is the parameter to control the trade-off between empirical error and the model complexity, and $s_{\ell_i} = +1 \text{ if } \textbf{y}_{i\ell} = 1, \text{ and } -1 \text{ otherwise } $.

For large scale problems, the above optimization problem is solved in the primal using trust region Newton method which requires the gradient and Hessian \cite{fan2008liblinear}.
The gradient of the objective function in Eq. \ref{objective} is given by 
\begin{displaymath}
\textbf{w}_{\ell} + 2C\textbf{X}_{I,:}^T(\textbf{X}_{I,:}\textbf{w}_{\ell} - \textbf{s}_{l_I})
\end{displaymath}
where $I$ is an index set of instances given by $\{i | 1-\textbf{s}_{l_i}\textbf{w}_l^T\textbf{x}_i > 0\}$.
The objective in Eq. \ref{objective} is not twice differentiable, and generalized Hessian is used which in closed form is given by
\begin{displaymath}
\mathcal{I} + 2C\textbf{X}_{I,:}^T\textbf{D}_{I, I}\textbf{X}_{I,:}
\end{displaymath}
where $\mathcal{I}$ is an identity matrix and $\textbf{D}_{I, I}$ is a diagonal matrix, both of dimensionality $|I|\times |I|$.
Solving the above optimization problem for a few hundred labels can be achieved by using Liblinear, but scaling them to XMC scenarios involving hundreds of thousands labels is a non-trivial task.
Direct usage of one-vs-rest for achieving parallel training can only use 32 or 64 cores (or more if available) on a single machine or node. Clearly, this is not sufficient for learning millions of weight vectors as in XMC. 
In order to tackle this issue, we develop DiSMEC which has two key components, (i) \textit{Double layer of parallelization} which enables learning thousands of $\textbf{w}_{\ell}$ in parallel and hence obtain significant speed-up for training in such scenarios, and, (ii) \textit{Model sparsity by restricting ambiguity} to control the growth in model sizes without sacrificing prediction accuracy.
These are explained in detail in the sections below.

\subsection{Double Layer of Parallelization}
We implement a two-layer parallelization architecture such that on the top level, labels are separated into batches of say 1,000, which are then sent to separate nodes each of which consists of 32 or 64 cores depending on cluster hardware.
On each node, parallel training of a batch of 1,000 labels is performed using openMP.
For a total of $L$ labels, it leads to $B=\lfloor\frac{L}{1000}\rfloor+1$ batches of labels.
For instance, if one has access to $M$ nodes each with 32 cores, then effectively $32 \times M$ labels can be trained in parallel. In our cluster, we had access to 32 such nodes, and approximately 1,000 labels could be trained in parallel.
Consequently, for WikiLSHTC-325K dataset, the computational problem boils down to solving 325 sequential binary problems of the form given in Equation \ref{objective}.
Using fast linear solvers such as Liblinear, a single binary problem consisting of approximately 1.5 million training instances in an approximately 1.5 million sparse feature vector space, can be solved in less than one minute.
As a result, the model for entire WikiLSHTC-325K dataset could be trained in approximately 3 hours on 1,000 cores. Using the same architecture, the model could be learnt in approximately 15 minutes on the Wikipedia-31K dataset consisting of 31,000 labels.
Note that the above architecture is flexible based on the number of cores available per node and number of total nodes that are available on the cluster hardware.

In addition to scalable and parallel training, learning models in batches of labels has another advantage - the resulting weight matrices are stored as individual blocks, each consisting of weight vectors for the number of labels in the batch, e.g., 1000 in our case. As we will describe later, these block matrices can be exploited for distributed prediction to achieve real-time prediction which is close to the performance of tree-based architecture such as FastXML.
The proposed DiSMEC learning framework is described in the algorithmic format in Algorithm \ref{algorithm:alg1}.

For efficient implementation, we avoid the need to replicate the input data $\textbf{X}$ for individual binary problem. This is done by creating data-structures such that $\textbf{X}$ is loaded in the main memory and shared to avoid any replication across $\mathcal{T}_l$, and only the sign vectors $\textbf{s}_l$ are maintained for individual labels. 
A naive data replication approach which stores the input data $\textbf{X}$ separately for each binary problem cannot scale to large-scale scenarios encountered in XMC.

\begin{algorithm}[!h]
\renewcommand{\algorithmicrequire}{\textbf{Input:}}
\renewcommand{\algorithmicensure}{\textbf{Output:}}
\begin{algorithmic}[1]
\Require{Training data $\mathcal{T} = \{(\textbf{x}_1,\textbf{y}_1) \ldots (\textbf{x}_n,\textbf{y}_n) \}$, input dimensionality $D$, label set $\{1 \ldots L\}$, $B=\lfloor\frac{L}{1000}\rfloor+1$  and $\Delta$}
\vspace{0.01in}
\Ensure{Learnt matrix $\textbf{W}_{D,L}$ in sparse format}
\vspace{0.01in}
\State Load single copy of input vectors $\textbf{X} = \{\textbf{x}_1 \ldots \textbf{x}_n \}$ in the main memory \Comment{\textit{Refactor data without replication}}
\vspace{0.01in}
\State Load binary sign vectors $\textbf{s}_l = \{+1,-1\}_{i=1}^n$ separately for each label in the main memory
\vspace{0.01in}
\For{$\{b=0; b < B; b++ \}$} \Comment{\textit{1st parallelization}}
\vspace{0.01in}
        \State \#pragma omp parallel for private($\ell$)  \Comment{\textit{2nd parallelization}}
        \vspace{0.01in}
       \For{$\{l=b\times1000; l \leq (b + 1)\times1000; l++ \}$}
       \vspace{0.01in}
\State Using $(\textbf{X}, \textbf{s}_l)$, train weight vector $\textbf{w}_\ell$ on a single core
\vspace{0.01in}
\State Prune ambiguous weights in $\textbf{w}_{\ell}$ \Comment{Model reduction}
\vspace{0.01in}
\EndFor
\vspace{0.01in}
\State \Return $\textbf{W}_{D,1000}$  \Comment{\textit{Learnt matrix for a batch on one node}}
\vspace{0.01in}
\EndFor
\vspace{0.01in}
\State \Return $\textbf{W}_{D,L}$ \Comment{\textit{Learnt matrix from all the nodes }}
\vspace{0.01in}
\end{algorithmic}
\caption{DiSMEC - Distributed Sparse Machines for Extreme Classification}
\label{algorithm:alg1}
\end{algorithm}

\begin{table*}[ht!]
\centering
\scalebox{0.9}{
\begin{tabular}{|c|c|c|c|c|c|c|}
\hline
  \textbf{Dataset} &  \# \textbf{Training } &  \# \textbf{Test } & \# \textbf{Categories} &  \# \textbf{Features} & \multicolumn{2}{c|}{\textbf{}}\\
  \cline{6-7}
  &&&& & \textbf{APpL} & \textbf{ALpP} \\
\hline
\textbf{Amazon-13K} &  1,186,239  & 306,782  & \textbf{13,330} & 203,882 & 448.5 & 5.04 \\
\textbf{Amazon-14K} &  4,398,050  & 1,099,725  & \textbf{14,588} & 597,540 & 1330.1 & 3.53 \\
\textbf{Wikipedia-31K} &  14,146  & 6,616  & \textbf{30,938} & 101,938 & 8.5 & 18.6 \\
\textbf{Delicious-200K} &  196,606  & 100,095  & \textbf{205,443} & 1,123,497 & 72.3  & 75.5 \\
\textbf{WikiLSHTC-325K} &  1,778,351 & 587,084  & \textbf{325,056} & 1,617,899 & 17.4 & 3.2 \\
\textbf{Wikipedia-500K} & 1,813,391 & 783,743  &  \textbf{501,070} &  2,381,304 & 24.7 & 4.7 \\
\textbf{Amazon-670K} & 490,499 & 153,025 & \textbf{670,091}  & 135909 & 3.9 & 5.4 \\
\hline
\end{tabular}
}
\caption{\footnotesize Multi-label datasets taken from the Extreme Classification Repository. APpL and ALpP represent average points per label and average labels per point respectively.}
\label{tbl:datasets}
\end{table*}
\subsection{Model sparsity via restricted ambiguity}

The method in the previous section solves the computational challenge of learning hundreds of thousand weight vectors, but still the models can aggregate upto 1 TB for large datasets such as WikiLSHTC-325K.
This is due to the use of $\ell_2$-regularization which squares the weight values encourages the presence of small weight values in the neighborhood of 0.
For the purpose of easier visualization, the distribution for the 3 Billion weights learnt for Wikipedia-31K dataset is shown in Figure \ref{fig:distorg}. It shows that a large fraction of these weights lie close to 0 (Y-axis is shown as a power of 10).
In fact, approximately 96\% of the weights lie in the interval $[-0.01, 0.01]$.
Similarly, for WikiLSHTC-325K dataset consisting of approximately 450 Billion ($325,000 \times 1,500,000$) weights, around 99.5\% weights lie in the interval $[-0.01, 0.01]$.
Storing all 450 Billion weights leads to model sizes of 870 GB as reported in the recent work \cite{yenpd}.

Let $\textbf{W}_{\Delta} = \{\textbf{W}_{d, \ell}, |\textbf{W}_{d,\ell}| < \Delta, 1 \leq d \leq D, 1 \leq \ell \leq L \}$.
For small values of $\Delta$, the weights values in $\textbf{W}_{\Delta}$, (i) represent the ambiguous weights which carry very little discriminative information of distinguishing one label against another, (ii) incur for large fraction of entries in the learnt matrix $\textbf{W}$ (96\% for Wikipdia-31K, and 99.5\% for WikiLSHTC-325K) (iii) occupy enormous disk-space rendering the learnt model useless for prediction purposes.
In principle, $\Delta$ can be thought of as an ambiguity control hyper-parameter of our DiSMEC algorithm. It controls the trade-off between model size and prediction speed on one hand, and reproducing the exact $\ell_2$-regularized model on the other hand. However, in practice, we fixed $\Delta = 0.01$, and it was observed to yield good performance. 

We therefore propose to set weights in $\textbf{W}_{\Delta}$ to 0 instead of storing them in the final model. 
For Wikipedia-31K dataset, the distribution weights after pruning with $\Delta = 0.01$ (shown in Figure \ref{fig:distprune}) leads to reduction in model size from 30GB to 500MB.
For WikiLSHTC-325K datasets, by setting $\Delta = 0.01$, the final model could be stored in 3GB instead of 870GB, which is around 3 orders of magnitude reduction.
It is also important to note that sparse models can also be achieved by using $\ell_1$-regularization but leads to under-fitting and worse prediction performance compared to using $\ell_2$ regularization followed by the pruning step, which is also verified in our experiments.
This pruning of weights can also be seen as learning the existence of only meaningful connections in two layered shallow network. 
Recently, compact model size has generated considerable interest in deep learning community on learning compact neural networks for embedded systems, low power devices and smartphones \cite{han2015learning, romero2014fitnets}.

\subsubsection{Impact on prediction speed and accuracy}
The drastic reduction in model size by pruning spurious weights enables easier access to models for making fast predictions.
The models for batches of labels (computed in step 9 of Algorithm \ref{algorithm:alg1}) $\textbf{W}_{D,1000}^1 \ldots \textbf{W}_{D,1000}^B$ can be stored in distributed manner, which can invoked in parallel for computing the inner product for predicting a test instance in an online manner.
The compact models coupled with distributed computations enable the application of DiSMEC framework making real-time predictions in XMC tasks such as recommendation systems and personalized search.

The pruning step with $\Delta=0.01$ has no adverse impact on prediction accuracy of the DiSMEC compared $\Delta=0$. This is because most the pruned weights are ambiguous weights in the neighborhood of 0, and have no significant discriminative power in distinguishing one label from another.
As a result, we obtain significant reduction in model size without any sacrifice in terms of generalization error.

\begin{table*}[ht!]
\centering
\scalebox{0.8}{
\begin{tabular}{|l||c||c|c|c||c|c|c||c|c|}
\hline
\multirow{2}{*}{Dataset} & \multicolumn{1}{c||}{Proposed~approach} & \multicolumn{3}{c||}{Embedding~based~approaches} & \multicolumn{3}{c||}{Tree~based~approaches} &  \multicolumn{2}{c|}{Sparsity~inducing~approaches}  \\
\cline{2-10}
&  \texttt{DiSMEC} & \texttt{SLEEC} & \texttt{LEML} & \texttt{RobustXML} &  \texttt{Fast-XML} & \texttt{LPSR-NB} & \texttt{PLT}  & \texttt{PD-Sparse} & \texttt{L1-SVM} \\
\hline
\textbf{Amazon-13K} &&&&& & & & & \\
\textit{P@1} &  \textbf{93.4}  & 90.4 &  78.2 & 88.4 & 92.9  & 75.1 & 91.4 & 91.1 &  91.8\\
\textit{P@3} & \textbf{79.1} & 75.8 & 65.4 & 74.6 & 77.5 & 60.2 & 75.8  & 76.4 &  77.8  \\
\textit{P@5} & \textbf{64.1} & 61.3 & 55.7 & 60.6 & 62.5 & 57.3 & 61.0 & 63.0.8 & 62.9 \\

\textbf{Amazon-14K} &&&&& & & & & \\
\textit{P@1} & \textbf{91.0}  & 80.3 &  75.2 & 83.2 & 90.3 & 74.2 & 86.4 & 88.4 & 88.2\\
\textit{P@3} & \textbf{70.3} & 67.2 & 62.5 & 66.4 & 70.1  & 55.7 & 65.2 & 68.1 & 67.6\\
\textit{P@5} & \textbf{55.9}  & 50.6 &  40.8 & 52.3 & 55.4  & 44.3 & 50.7 & 50.5  &  51.2\\

\textbf{Wikipedia-31K} &&&&& & & & & \\
\textit{P@1} &  85.2 & 85.5 &  73.5 & \textbf{85.5} & 82.5  & 72.7 & 84.3 & 73.8 & 83.2\\
\textit{P@3} & \textbf{74.6} & 73.6 & 62.3 & 74.0 & 66.6 & 58.5 & 72.3 & 60.9 & 72.1\\
\textit{P@5} & \textbf{65.9} & 63.1 & 54.3 & 63.8 & 56.7 & 49.4 & 62.7 & 50.4& 63.7\\

\textbf{Delicious-200k} &&&&& & & & & \\
\textit{P@1} &  45.5  & \textbf{47.0} &  40.3 & 45.0 & 42.8  & 18.6 & 45.3 & 41.2 & 42.1\\
\textit{P@3} & 38.7 & \textbf{41.6} & 37.7 &40.0 & 38.7 & 15.4 &  38.9 & 35.3 & 34.8\\
\textit{P@5} & 35.5 & \textbf{38.8} & 36.6 & 38.0 & 36.3 & 14.0 & 35.8 & 31.2 & 30.4\\

\textbf{WikiLSHTC-325K} &&&&& & & & & \\
\textit{P@1} &  \textbf{64.4}  & 55.5 & 19.8 & 53.5 & 49.3  & 27.4 & 45.6 & 58.2 & 60.6\\
\textit{P@3} & \textbf{42.5} & 33.8 & 11.4 & 31.8 & 32.7 & 16.4 &29.1 & 36.3 & 38.6\\
\textit{P@5} & \textbf{31.5} & 24.0 & 8.4 & 29.9 & 24.0 & 12.0 & 21.9  & 28.7 & 28.5\\

\textbf{Wiki-500K} &&&&& & & & & \\
\textit{P@1} &  \textbf{70.2}  & 48.2 & 41.3 & - & 54.1  & 38.2  & 51.5  &-&65.3\\
\textit{P@3} & \textbf{50.6} & 29.4 &  30.1 &- & 35.5 & 29.3 & 35.7 &-&46.1\\
\textit{P@5} & \textbf{39.7} & 21.2 & 19.8 & -& 26.2 & 18.7 & 27.7 &-&35.3\\

\textbf{Amazon-670K} &&&&& & & & & \\
\textit{P@1} &  \textbf{44.7}  & 35.0 &  8.1 & 31.0 & 33.3  & 28.6 & 36.6  &- &39.8\\
\textit{P@3} & \textbf{39.7} & 31.2 & 6.8 & 28.0 & 29.3 & 24.9 & 32.1 & -& 34.3\\
\textit{P@5} & \textbf{36.1} & 28.5 & 6.0 & 24.0 &  26.1 & 22.3  & 28.8 &- & 30.1 \\

\hline
\end{tabular}
}
\caption{Comparison of Precision@k for k=1,3 and 5, the entries marked - denote instances when the method could not scale to that particular dataset}
\label{tbl:MiMa}
\end{table*}

\section{Experimental evaluation}
\subsection{Dataset description}
We present elaborate results on publicly available datasets from the Extreme Classification repository \footnote{\url{http://research.microsoft.com/en-us/um/people/manik/downloads/XC/XMLRepository.html}}.
These datasets are curated from sources such as Wikipedia, Amazon and Delicious \cite{partalas2015lshtc, mcauley2013hidden, wetzkeranalyzing}.
Wikipedia pages have labels tagged at the bottom of every page, Amazon is a product to product recommendation dataset, and Delicious is a dataset from social bookmarking domain.

The number of labels for the datasets used in our empirical evaluation vary from 13,330 to 670,091.
The number of training points and features also range upto 4.3 million and 2.3 million respectively.
The detailed statistics for individual datasets are shown in Table \ref{tbl:datasets}.
The split between train and test set are kept exactly the same as given on XML repository page.
The datasets appear in the sparse LibSVM format with labels mapped as positive integers. 
Furthermore, no side or additional information from any external sources was added to gain any advantage.

\subsection{Evaluation Metrics}
We use precision at $k$, denoted P@k, and normalized Discounted Cumulative Gain, denoted nDCG@k as the metrics for comparison. 
These are commonly used metrics in extreme classification, and have been reported by most state-of-the-art methods, and bench-marked in the extreme classification repository\cite{Agrawal13, bhatia2015sparse, prabhu2014fastxml, xurobust}.
For a label space of dimensionality $L$, a true label vector $\textbf{y} \in \{0,1\}^L$ and predicted label vector $\hat{\textbf{y}} \in \mathbb{R}^L$, the metrics are defined as follows :
\begin{itemize}
\item P@k := $ \sum_{l \in rank_k{(\hat{\textbf{y}})}}{\textbf{y}_l}$

\item nDCG@k := $\frac{1}{k} \sum_{l \in rank_k{(\hat{\textbf{y}})}}{\frac{\textbf{y}_l}{\log(l+1)}}$
\end{itemize}
where nDCG@k := $\frac{\text{DCG@k}}{\sum_{l=1}^{\min(k, ||\textbf{y}||_0)}{\frac{1}{\log(l+1)}}}$, $rank_k(\textbf{y})$ returns the $k$ largest indices of $\textbf{y}$ ranked in descending order, and $||\textbf{y}||_0$ returns the 0-norm of the true-label vector. Note that, unlike P@k, nDCG@k takes into account the ranking of the correctly predicted labels. For instance, if there is only one of the five labels that is correctly predicted, then P@5 gives the same score if the correctly predicted label is at rank 1 or rank 5. On the other hand, nDCG@5 gives it a higher score if it predicted at rank 1, and the lowest non-zero score if it is predicted at rank 5.

It may be noted that standard performance measure for multi-label classification such as Hamming loss, Micro-F1 and Macro-F1 are not wuite suited for XMC setting because of the following reasons, Firstly, (i) in most real world XMC applications such as document tagging and recommendation systems, $k$ slots are available for making prediction and it is more important to make relevant predictions for these top $k$ slots, rather attempting to predict all the labels, which may be much more than $k$, and secondly (ii) in XMC settings, no expert can verify the entire label set for each training or test instance. As a result, there could be many missing labels per instance. In such scenarios, measures such as Hamming loss or Micro-F1 and Macro F1 may not reflect the true underlying performance.
\subsection{Methods for comparison}
\begin{figure*}[ht]
\centering

\subfloat[Amazon-670K]{%
  \includegraphics[width=0.43\textwidth]{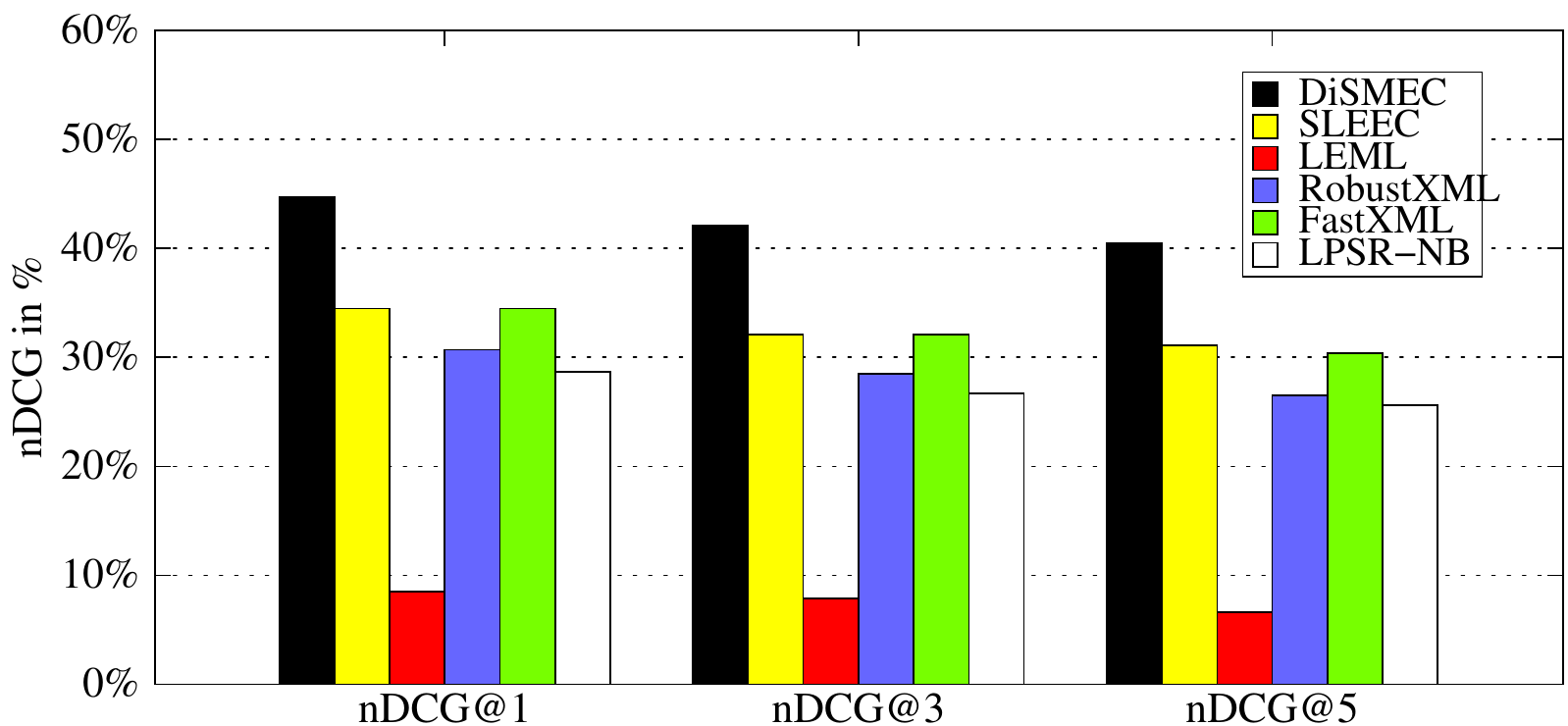}%
  \label{amazonbar}%
}
~
\subfloat[Delicious-200K]{%
  \includegraphics[width=0.43\textwidth]{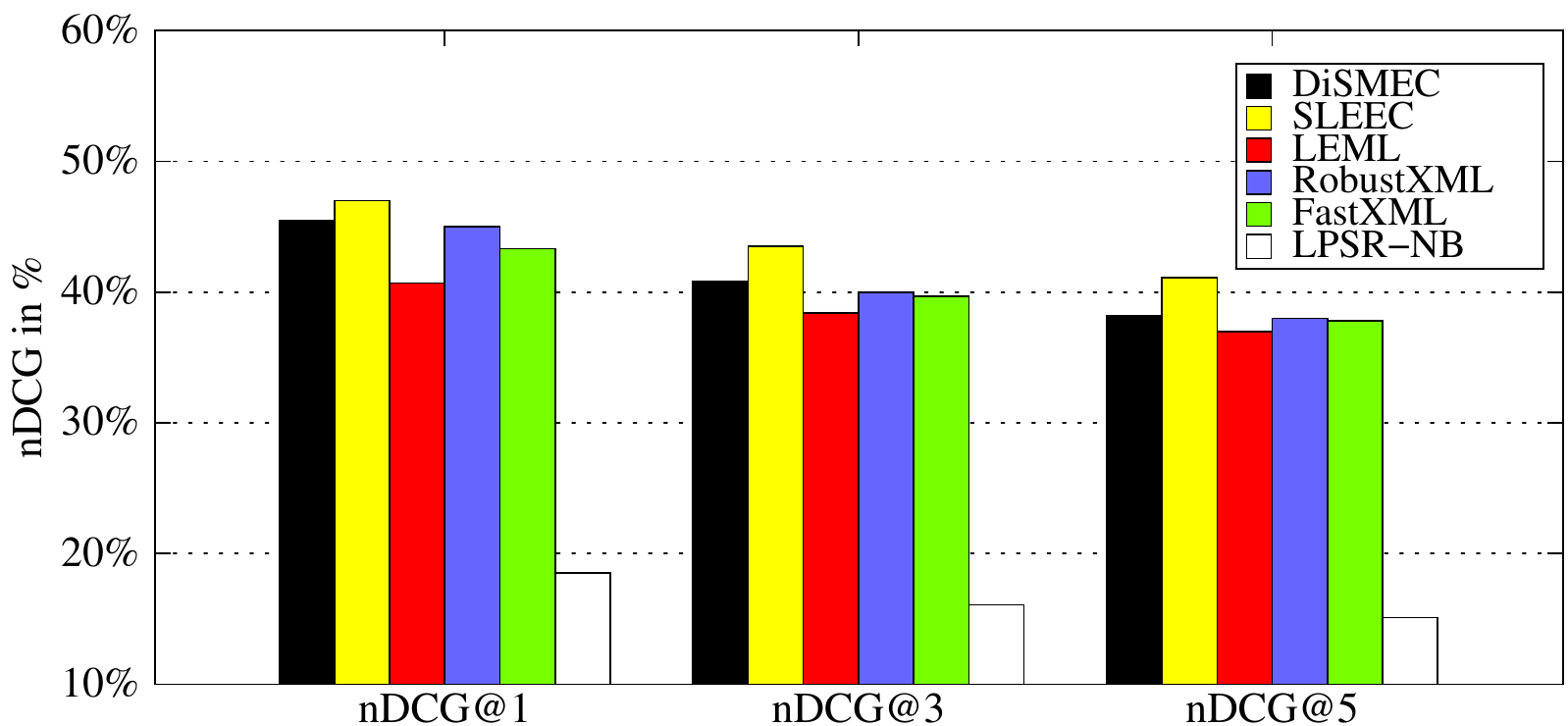}%
  \label{deliciousbar}%
}

\subfloat[WikiLSHTC-325K]{%
  \includegraphics[width=0.43\textwidth]{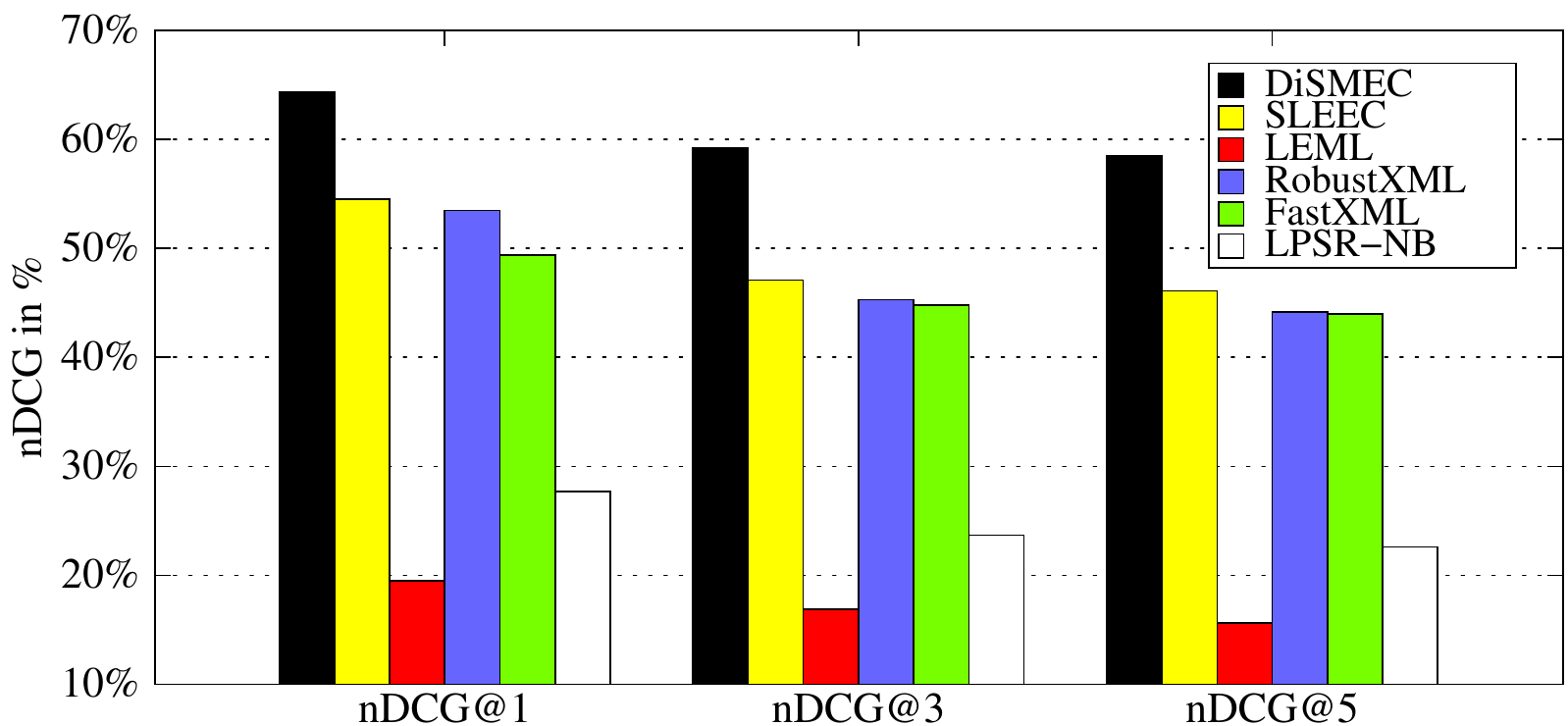}%
  \label{wiki325bar}%
}
~
\subfloat[Wikipedia-31K]{%
  \includegraphics[width=0.43\textwidth]{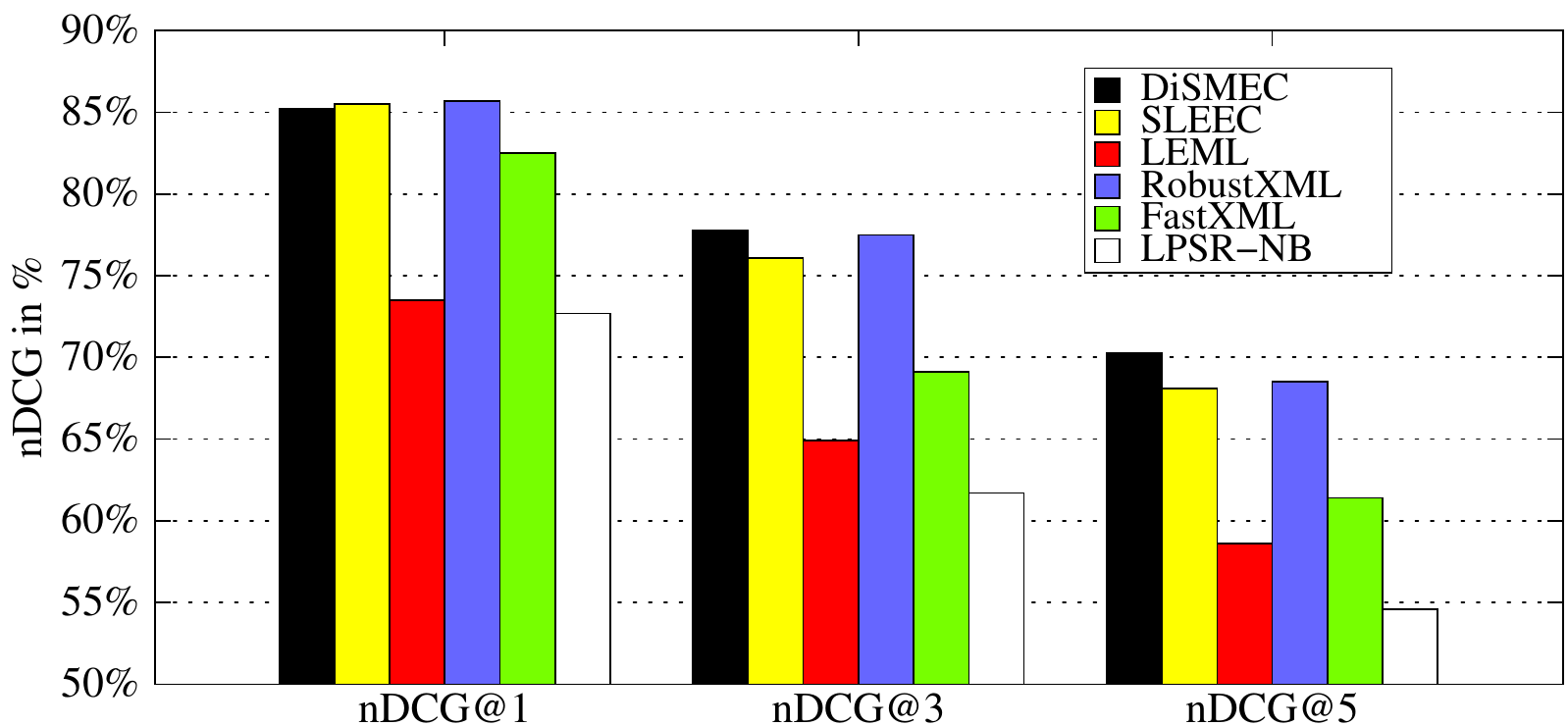}%
  \label{wiki10bar}%
}

\caption{nDCG@k for k=1, 3 and 5}
\label{fig:ndcg}
\end{figure*}

DiSMEC was implemented in C++ on 64-bit Linux system over Liblinear package using openMP for parallelization. The code will be publicly available shortly.
We compare \texttt{DiSMEC} as proposed in Algorithm \ref{algorithm:alg1} against eight baselines, some of which are state-of-the-art methods deployed in industrial settings.
These are listed below:
\begin{itemize}
\item Embedding-based methods which project the label matrix into a low-dimensional sub-space :
\begin{enumerate}
\item \texttt{SLEEC} \cite{bhatia2015sparse} - This is the state-of-the-art approach for learning sparse \textit{local} embeddings in extreme classification.
\item \texttt{LEML} \cite{yu2014large}- This method proposes a global embedding of the label space, and may not be suitable when there exists a large fraction of tail labels when the low dimension assumption is violated.
\item \texttt{RobustXML} \cite{xurobust} - This approach also makes a low-dimensional embedding approach but assumes the tail labels to be outliers and treats them separately in the overall optimization problem.
\end{enumerate}

\item Tree-based methods are designed for faster prediction by cutting the search space of the potential labels. Since wrong predictions at top level cannot be recovered, these methods typically trade-off prediction accuracy with prediction speed.
\begin{enumerate}
\item \texttt{Fast-XML} \cite{prabhu2014fastxml} - This is a tree-based method which performs partitioning in the feature space for faster prediction. The objective function directly optimizes an nDCG based ranking loss function.
\item \texttt{LPSR-NB} \cite{weston2013label} - This method learns a hierarchy of labels in addition to learning the classifier for the entire label set. Since computational cost for both the steps can be extremely high in the XMC scenarios, it was only possible to train Naive Bayes classifier as the base classifier in the ensemble.
\item \texttt{PLT} \cite{jasinska} - In this recently proposed approach, the goal is to maximize F-measure, which is achieved by using Expected Utility Maximization framework by tuning a threshold for label probability estimates to separate the positive labels from the negative ones.
\end{enumerate}
\item Other baselines 
\begin{enumerate}
\item \texttt{L1-SVM} \cite{fan2008liblinear} - In this method, $\ell_1$-regularization is used along with binary one-vs-rest loss. Though it gives much sparser solutions, but its models typically underfit the data leading to degradation in prediction accuracy. 
\item \texttt{PD-Sparse} \cite{yenpd} - This is a very recently proposed approach which uses $\ell_1$ regularization along with multi-class loss instead of binary one-vs-rest loss. As a result, it needs to store the intermediary weight vectors during optimization, and hence did not scale to large datasets.
\end{enumerate}
\end{itemize}
We do not compare explicitly against the random forest approach as proposed in \cite{Agrawal13}, as it has already been shown that SLECC and FastXML perform better than the random forest classifier.
For DiSMEC, the hyper-parameter $C$ was set on a validation set which was extracted from the training set, and $\Delta$ was set to 0.01 for all the datasets.
For all other approaches, hyper-parameters were set as suggested in the software, or were set to default values due to limitations of dataset or computational resources. For instance, the number of learners in SLEEC for Wikipedia-500K dataset was set to 5 instead of 10 for the computation to finish within few days.
\section{Results}
\subsection{Prediction Accuracy}

The results for precision@k for k=1,3, and 5 and nDCG@k for k=1,3, and 5 are shown in Table \ref{tbl:MiMa} and Figure \ref{fig:ndcg} respectively. In the interest of space, the nDCG results are shown only for four datasets.
It is clear that DiSMEC can out-perform state-of-the-art baselines on most datasets. 
For Amazon-670K, WikiLSHTC-325K, and Wiki-500K datasets, the improvements in precision@k over SLEEC are of the order of 10\% points in absolute terms. 
On a relative scale, the gains achieved by DiSMEC are in the range of 20-to-25\% points over these methods.
For these datasets, since the label set is extremely large and diverse, exhibits a power-law phenomena, and there are very few labels per instance, this suggests that the label correlations and (local) low-rank are virtually non-existent in these datasets.
Strong improvements gained by a scalable one-vs-rest scheme DiSMEC, over the embedding based methods, confirms this intuition further.

On the other hand, since Delicious-200k dataset has a relatively larger average number of labels per training instance (shown in last column of Table \ref{tbl:datasets}). It is likely that this leads to much stronger label correlations in this dataset compared to the other datasets, and hence embedding-based methods such as SLEEC can perform slightly better than DiSMEC.

As compared to FastXML, which is a leading tree-based approach, the gains in precision@k and nDCG@k are even substantial, of the order of 15\% for larger datasets. The tree-based methods which partition the label space or feature space compensate prediction accuracy for better prediction speed. DiSMEC performs model training in batches of labels and it also stores the models in a distributed manner. These can be evaluated in parallel when doing prediction  to achieve real-time speed and performance competitive to tree-based methods.

\begin{figure}[ht]
\centering

\subfloat[Distribution of weights for $l_1$-regularization]{%
  \includegraphics[width=0.5\textwidth]{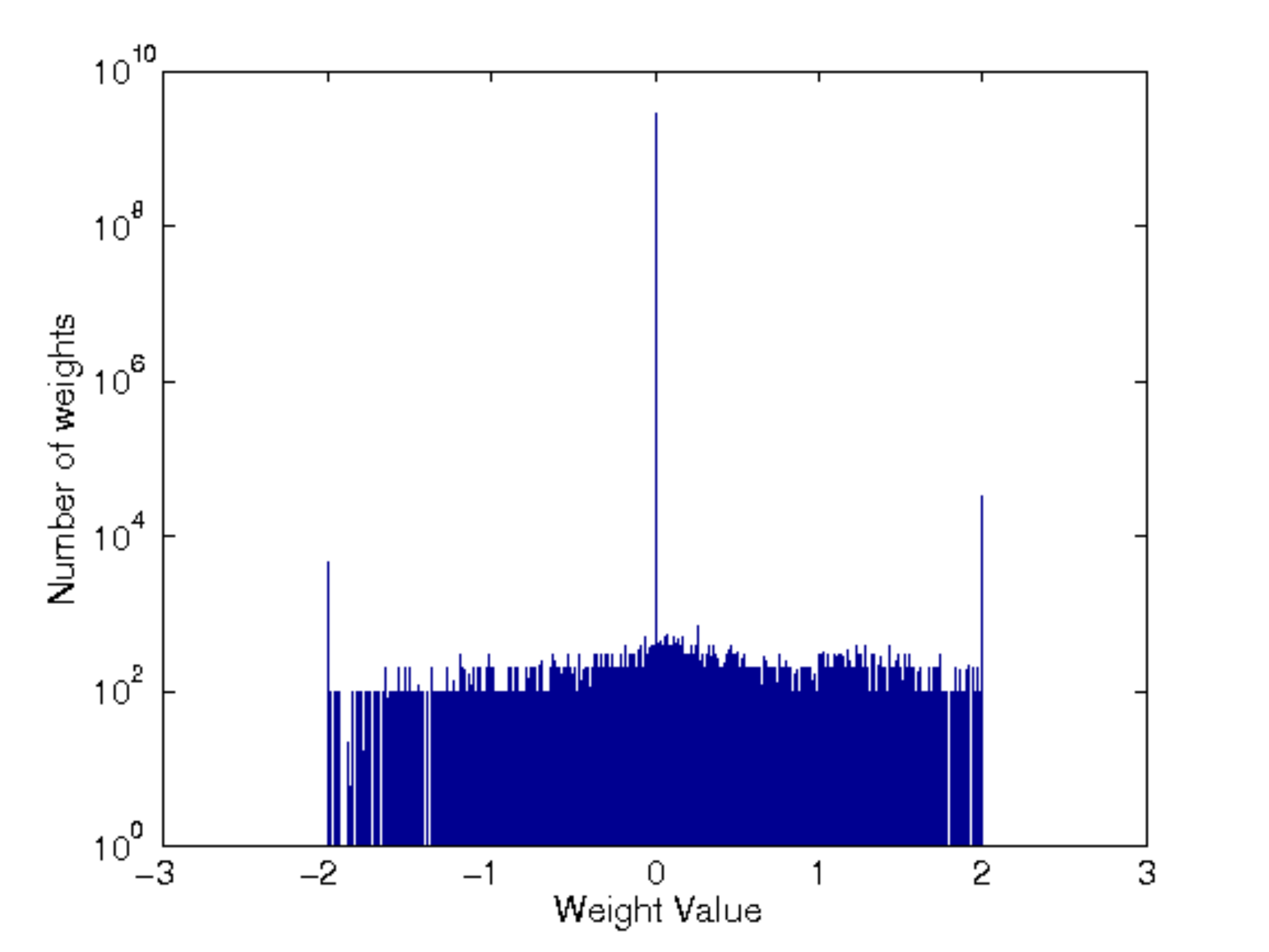}%
  \label{l1reg}%
}

\caption{Histogram plot depicting the distribution of learnt weights for Wikipedia-31K dataset by using $l_1$-regularization}
\end{figure}

DiSMEC framework which uses $l_2$-regularized SVM can also be used employed for $l_1$-regularization. In that case, one does not need to perform the explicit sparsity step (step 7 in Algorithm \ref{algorithm:alg1}) since the learning algorithm itself performs strong feature selection. For Wikipedia-31K datasets, the distribution of weights for $l_1$-regularization is shown in Figure \ref{l1reg}. Due to the very few non-zero features, the resulting model underfits the data and gives lower prediction performance compared to $l_2$-regularization. The advantage of this approach is that it can get even smaller models than the models learnt by default DiSMEC algorithm and still be competitive.
PD-Sparse which performs uses multi-class loss instead of binary one-vs-rest loss performs at par with \texttt{L1-SVM} for smaller datasets. However, it could not be scaled to the large datasets due to main memory constraints.

\begin{figure*}[ht]
\centering

\subfloat[Variation in precision@k]{%
  \includegraphics[width=0.4\textwidth]{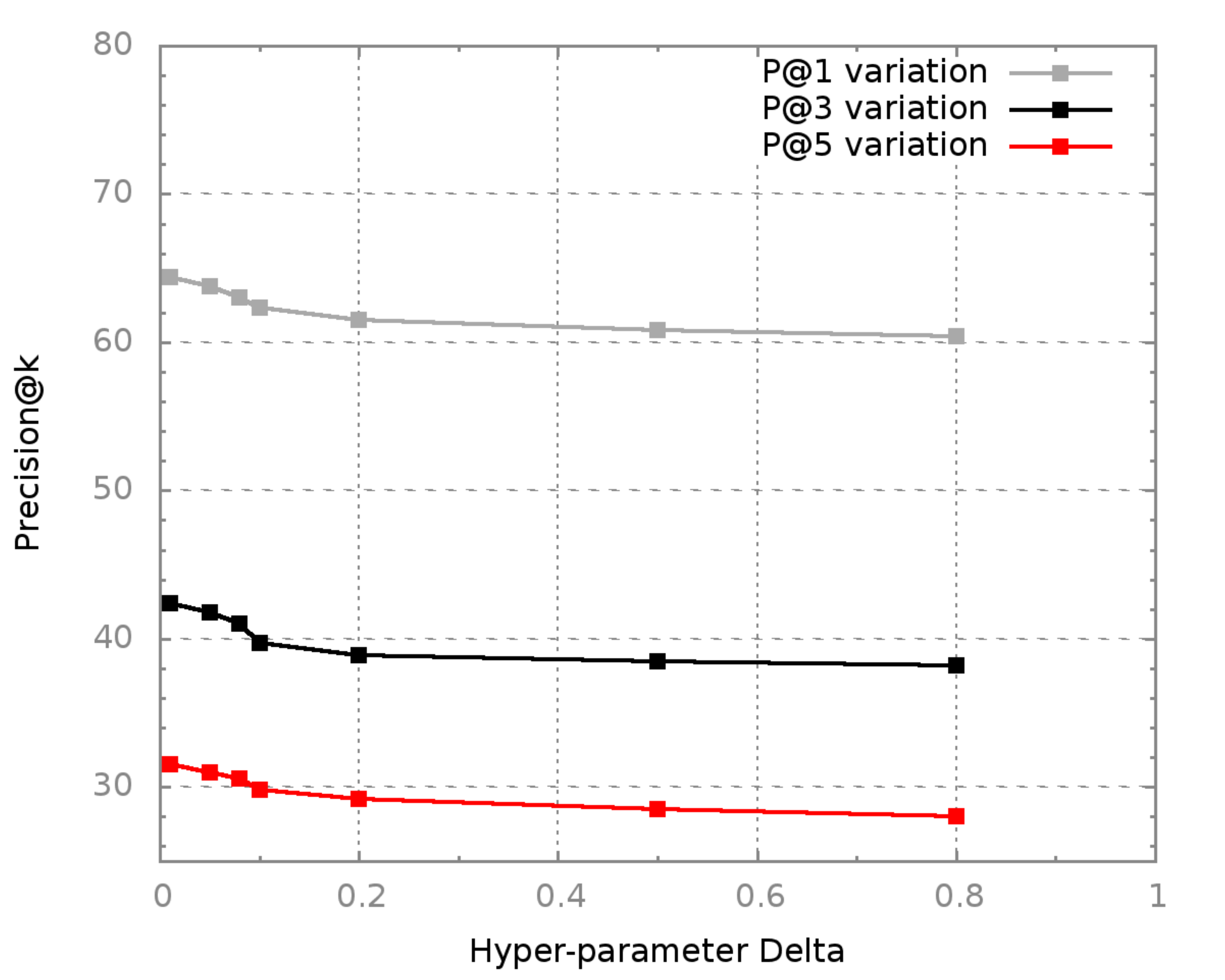}%
  \label{amazonbar}%
}
~
\subfloat[Variation in model size]{%
  \includegraphics[width=0.4\textwidth]{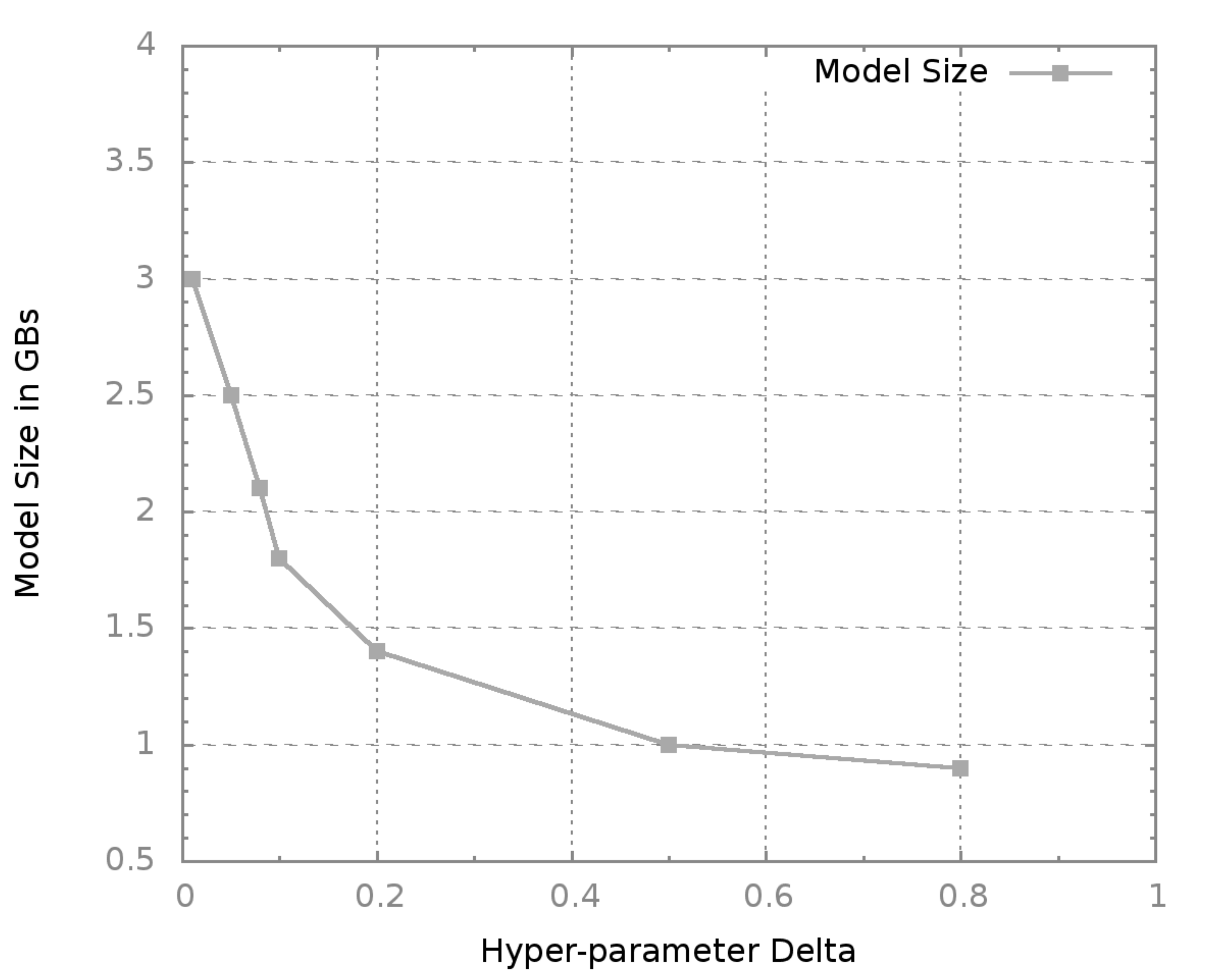}%
  \label{deliciousbar}%
}

\caption{Impact of $\Delta$ on precision@k and model size for WikiLSHTC-325K dataset}
\label{fig:delta}
\end{figure*}

\subsection{Model Size}
The proposed DiSMEC framework controls the model size by filtering out the billions of spurious parameters stored by ad-hoc usage of the off-the-shelf solvers.
By weeding out these ambiguous parameters (step 7 in Algorithm \ref{algorithm:alg1}), one can obtain model sizes which are three orders of magnitude smaller. 
In fact, for WikiLSHTC-325K dataset, the learnt model is 3GB instead of 870GB mentioned in the recent work (\cite{yenpd} Table 2). 
For the model size fixed at 3GB model size, SLEEC which aggregates an ensemble of learners gives only 52\% accuracy for precision@1 compared to only one learner of DiSMEC which gives 64.4\%.
This shows that model pruning for storing only meaningful information in the compact final model is a crucial part of DiSMEC learning framework in order to make meaningful predictions.
The impact of $\Delta$ on the model size for WikiLSHTC-325 dataset, is also studied later.

\subsection{Training and prediction Complexities}
The double layer of parallelization in DiSMEC framework enables flexible access to computing units available to the algorithm. It can reduce the training time from several weeks (as mentioned in \cite{yenpd} Table 1 which use sequential training provided by off-the-shelf solvers) to within few hours of training time on datasets consisting hundreds of thousand labels. 
For Wikipedia-31K, DiSMEC trained in approximately 10 minutes on 300 cores.
For WikiLSHTC-325K, DISMEC took approximately 6 hours on 400 cores and 3 hours on 1,000 cores. Furthermore, the architecture is very flexible and can exploit as many free cores as are available, potentially equalling the number of number of labels. On other hand, PD-Sparse which multi-class loss, cannot be inherently parallelized and required hundreds of GBs of main memory while training and 2 days of training.

The batch training of models (step 3, Algorithm \ref{algorithm:alg1}) in the parallel learning architecture of DiSMEC enables distributed storage of models. 
As a result, the vector-matrix product for the new test instance can be in parallel for the $B$ block matrices corresponding to each batch (step 9, Algorithm \ref{algorithm:alg1}).
Furthermore, due to sparsity both in the input test instance and the learnt model from the sparsity inducing step, the dot product can be evaluated extremely efficiently.
For WikiLSHTC-325K, DiSMEC took 3 milliseconds per test instance which is thrice as fast as SLEEC, 1,000 times faster than LEML and also comparable to tree-based methods such as FastXML which takes 0.5 milliseconds per test instance.
This prediction speed achieved by the distributed prediction mechanism in DiSMEC is therefore well suited for making real-time applications of extreme classification in recommendation systems and ranking.

\subsection{Impact of $\Delta$ }

The hyper-parameter $\Delta >0$ in DiSMEC controls the trade-off between model size and prediction accuracy. 
For WikiLSHTC-325K dataset, the impact of variation of $\Delta$ on precision@k and model size is shown in Figure \ref{fig:delta}.
Setting $\Delta=0$ gives back the original model without the explicit sparsity induced in Step 7 of Algorithm \ref{algorithm:alg1}.
Setting $\Delta >0$ away from 0 gives sparser models since more of the weights in the learnt matrix are pruned. 
On the other hand, this leads to reduction in prediction accuracies compared to $\Delta=0.01$. 
The model could not be tested for the setting $\Delta=0$ due to extremely large size. 
For smaller datasets, it was observed that changing from $\Delta=0$ to $\Delta=0.01$ has no significant impact on prediction accuracy. For some cases, there was a marginal improvement in accuracy and for some others there was a marginal decline.
The parameter $\Delta$ essentially tunes the model behavior between the two extremes of $l_2$ and $l_1$ regularization.
\section{Conclusion and Future work}
In this work, we presented DiSMEC learning framework for extreme multi-label learning. By employing the doubly parallelized architecture and explicit model sparsity induction, we showed that a careful implementation of one-vs-rest mechanism can result in drastic improvements in prediction accuracies over state-of-the-art industrial machine learning systems. Furthermore, the proposed method can easily scale for problems involving hundreds of thousand labels, and the compact models learnt can be used for real-time applications of extreme classification such as recommendation systems and ranking.
The avenues for future work include, (i) focussing on performance of classifier on the coverage of tail labels which are normally under-represented in the training set, (ii) formalize the notion of degree correlation among labels, and using this to design a criterion to choose between embedding-based methods and one-vs-rest approaches and (iii) studying the impact of power-law exponent on the choice of methods.

\bibliographystyle{abbrv}
\bibliography{LSHTC-biblio}

\begin{thebibliography}{10}

\bibitem{Agrawal13}
R.~Agrawal, A.~Gupta, Y.~Prabhu, and M.~Varma.
\newblock Multi-label learning with millions of labels: Recommending advertiser
  bid phrases for web pages.
\newblock In {\em Proceedings of the International World Wide Web Conference},
  May 2013.

\bibitem{babbar2014power}
R.~Babbar, C.~Metzig, I.~Partalas, E.~Gaussier, and M.-R. Amini.
\newblock On power law distributions in large-scale taxonomies.
\newblock {\em ACM SIGKDD Explorations Newsletter}, 16(1):47--56, 2014.

\bibitem{babbar2013flat}
R.~Babbar, I.~Partalas, E.~Gaussier, and M.-R. Amini.
\newblock On flat versus hierarchical classification in large-scale taxonomies.
\newblock In {\em Advances in Neural Information Processing Systems}, pages
  1824--1832, 2013.

\bibitem{babbar2014re}
R.~Babbar, I.~Partalas, E.~Gaussier, and M.-r. Amini.
\newblock Re-ranking approach to classification in large-scale power-law
  distributed category systems.
\newblock In {\em Proceedings of the 37th international ACM SIGIR conference on
  Research \& development in information retrieval}, pages 1059--1062. ACM,
  2014.

\bibitem{Bengio10}
S.~Bengio, J.~Weston, and D.~Grangier.
\newblock Label embedding trees for large multi-class tasks.
\newblock In {\em Neural Information Processing Systems}, pages 163--171, 2010.

\bibitem{bhatia2015sparse}
K.~Bhatia, H.~Jain, P.~Kar, M.~Varma, and P.~Jain.
\newblock Sparse local embeddings for extreme multi-label classification.
\newblock In {\em Advances in Neural Information Processing Systems}, pages
  730--738, 2015.

\bibitem{bi2013efficient}
W.~Bi and J.~Kwok.
\newblock Efficient multi-label classification with many labels.
\newblock In {\em Proceedings of The 30th International Conference on Machine
  Learning}, pages 405--413, 2013.

\bibitem{chen2012feature}
Y.-N. Chen and H.-T. Lin.
\newblock Feature-aware label space dimension reduction for multi-label
  classification.
\newblock In {\em Advances in Neural Information Processing Systems}, pages
  1529--1537, 2012.

\bibitem{cisse2013robust}
M.~M. Cisse, N.~Usunier, T.~Artieres, and P.~Gallinari.
\newblock Robust bloom filters for large multilabel classification tasks.
\newblock In {\em Advances in Neural Information Processing Systems}, pages
  1851--1859, 2013.

\bibitem{deng2010does}
J.~Deng, A.~C. Berg, K.~Li, and L.~Fei-Fei.
\newblock What does classifying more than 10,000 image categories tell us?
\newblock In {\em Computer Vision--ECCV 2010}, pages 71--84. Springer, 2010.

\bibitem{fan2008liblinear}
R.-E. Fan, K.-W. Chang, C.-J. Hsieh, X.-R. Wang, and C.-J. Lin.
\newblock Liblinear: A library for large linear classification.
\newblock {\em The Journal of Machine Learning Research}, 9:1871--1874, 2008.

\bibitem{gopal2013recursive}
S.~Gopal and Y.~Yang.
\newblock Recursive regularization for large-scale classification with
  hierarchical and graphical dependencies.
\newblock In {\em Proceedings of the 19th ACM SIGKDD international conference
  on Knowledge discovery and data mining}, pages 257--265. ACM, 2013.

\bibitem{han2015learning}
S.~Han, J.~Pool, J.~Tran, and W.~Dally.
\newblock Learning both weights and connections for efficient neural network.
\newblock In {\em Advances in Neural Information Processing Systems}, pages
  1135--1143, 2015.

\bibitem{hsu2009multi}
D.~Hsu, S.~Kakade, J.~Langford, and T.~Zhang.
\newblock Multi-label prediction via compressed sensing.
\newblock In {\em Advances in neural information processing systems}, 2009.

\bibitem{jasinska}
K.~Jasinska, K.~Dembczynski, R.~Busa{-}Fekete, K.~Pfannschmidt, T.~Klerx, and
  E.~H{\"{u}}llermeier.
\newblock Extreme f-measure maximization using sparse probability estimates.
\newblock In {\em Proceedings of the 33nd International Conference on Machine
  Learning}, pages 1435--1444.

\bibitem{krizhevsky2012imagenet}
A.~Krizhevsky, I.~Sutskever, and G.~E. Hinton.
\newblock Imagenet classification with deep convolutional neural networks.
\newblock In {\em Advances in neural information processing systems}, pages
  1097--1105, 2012.

\bibitem{linmulti}
Z.~Lin, G.~Ding, M.~Hu, and J.~Wang.
\newblock Multi-label classification via feature-aware implicit label space
  encoding.
\newblock pages 325--333, 2014.

\bibitem{mcauley2013hidden}
J.~McAuley and J.~Leskovec.
\newblock Hidden factors and hidden topics: understanding rating dimensions
  with review text.
\newblock In {\em Proceedings of the 7th ACM conference on Recommender
  systems}, pages 165--172. ACM, 2013.

\bibitem{partalas2015lshtc}
I.~Partalas, A.~Kosmopoulos, N.~Baskiotis, T.~Artieres, G.~Paliouras,
  E.~Gaussier, I.~Androutsopoulos, M.-R. Amini, and P.~Galinari.
\newblock Lshtc: A benchmark for large-scale text classification.
\newblock {\em arXiv preprint arXiv:1503.08581}, 2015.

\bibitem{Prabhu14}
Y.~Prabhu and M.~Varma.
\newblock Fastxml: A fast, accurate and stable tree-classifier for extreme
  multi-label learning.
\newblock In {\em Proceedings of the ACM SIGKDD}, August 2014.

\bibitem{prabhu2014fastxml}
Y.~Prabhu and M.~Varma.
\newblock Fastxml: A fast, accurate and stable tree-classifier for extreme
  multi-label learning.
\newblock In {\em Proceedings of the 20th ACM SIGKDD international conference
  on Knowledge discovery and data mining}, pages 263--272. ACM, 2014.

\bibitem{rifkin2004defense}
R.~Rifkin and A.~Klautau.
\newblock In defense of one-vs-all classification.
\newblock {\em Journal of machine learning research}, 5(Jan):101--141, 2004.

\bibitem{romero2014fitnets}
A.~Romero, N.~Ballas, S.~E. Kahou, A.~Chassang, C.~Gatta, and Y.~Bengio.
\newblock Fitnets: Hints for thin deep nets.
\newblock {\em arXiv preprint arXiv:1412.6550}, 2014.

\bibitem{shen2011item}
D.~Shen, J.~D. Ruvini, M.~Somaiya, and N.~Sundaresan.
\newblock Item categorization in the e-commerce domain.
\newblock In {\em Proceedings of the 20th ACM international conference on
  Information and knowledge management}, pages 1921--1924. ACM, 2011.

\bibitem{tai2012multilabel}
F.~Tai and H.-T. Lin.
\newblock Multilabel classification with principal label space transformation.
\newblock {\em Neural Computation}, pages 2508--2542, 2012.

\bibitem{weston2011wsabie}
J.~Weston, S.~Bengio, and N.~Usunier.
\newblock Wsabie: Scaling up to large vocabulary image annotation.
\newblock 2011.

\bibitem{weston2013label}
J.~Weston, A.~Makadia, and H.~Yee.
\newblock Label partitioning for sublinear ranking.
\newblock In {\em Proceedings of The 30th International Conference on Machine
  Learning}, pages 181--189, 2013.

\bibitem{wetzkeranalyzing}
R.~Wetzker, C.~Zimmermann, and C.~Bauckhage.
\newblock Analyzing social bookmarking systems: A del. icio. us cookbook.

\bibitem{xurobust}
C.~Xu, D.~Tao, and C.~Xu.
\newblock Robust extreme multi-label learning.
\newblock In {\em Proceedings of the ACM SIGKDD international conference on
  Knowledge discovery and data mining}, 2016.

\bibitem{yenpd}
I.~E. Yen, X.~Huang, P.~Ravikumar, K.~Zhong, and I.~S. Dhillon.
\newblock Pd-sparse : A primal and dual sparse approach to extreme multiclass
  and multilabel classification.
\newblock In {\em Proceedings of the 33nd International Conference on Machine
  Learning}, 2016.

\bibitem{yu2014large}
H.-F. Yu, P.~Jain, P.~Kar, and I.~Dhillon.
\newblock Large-scale multi-label learning with missing labels.
\newblock In {\em Proceedings of The 31st International Conference on Machine
  Learning}, pages 593--601, 2014.

\bibitem{yuan2012recent}
G.-X. Yuan, C.-H. Ho, and C.-J. Lin.
\newblock Recent advances of large-scale linear classification.
\newblock {\em Proceedings of the IEEE}, 100(9):2584--2603, 2012.

\bibitem{zhangmulti}
Y.~Zhang and J.~Schneider.
\newblock Multi-label output codes using canonical correlation analysis.
\newblock pages 873--882, 2011.

\end{thebibliography}
\end{document}